\documentclass[conference]{IEEEtran}
\usepackage{amsmath,amsfonts}
\usepackage[numbers]{natbib}
\usepackage{array}
\usepackage[caption=false,font=normalsize,labelfont=sf,textfont=sf]{subfig}
\usepackage{textcomp}
\usepackage{stfloats}
\usepackage{url}
\usepackage{float}
\usepackage{verbatim}
\usepackage{graphicx}
\usepackage{blindtext}
\usepackage{capt-of}
\usepackage{bbding}
\usepackage[table,xcdraw]{xcolor}
\usepackage[normalem]{ulem} 
\usepackage{multicol}
\usepackage{caption}
\usepackage[bookmarks=true]{hyperref}
\begin{document}

\title{
Dexterous Cable Manipulation: Taxonomy, Multi-Fingered Hand Design, and Long-Horizon Manipulation
}

\author{Sun Zhaole$^{1}$, Xiao Gao$^{2}$, Xiaofeng Mao$^{1}$, Jihong Zhu$^{3}$, Aude Billard$^{2}$, and Robert B. Fisher$^{1}$

\\
$^1$University of Edinburgh~~~
$^2$EPFL~~~
$^3$University of York~~~
\\
Project Web: https://sites.google.com/view/dexterous-cable-manipulation/home

}

\twocolumn[{%
\renewcommand\twocolumn[1][]{#1}%
\maketitle
\includegraphics[width=\linewidth]{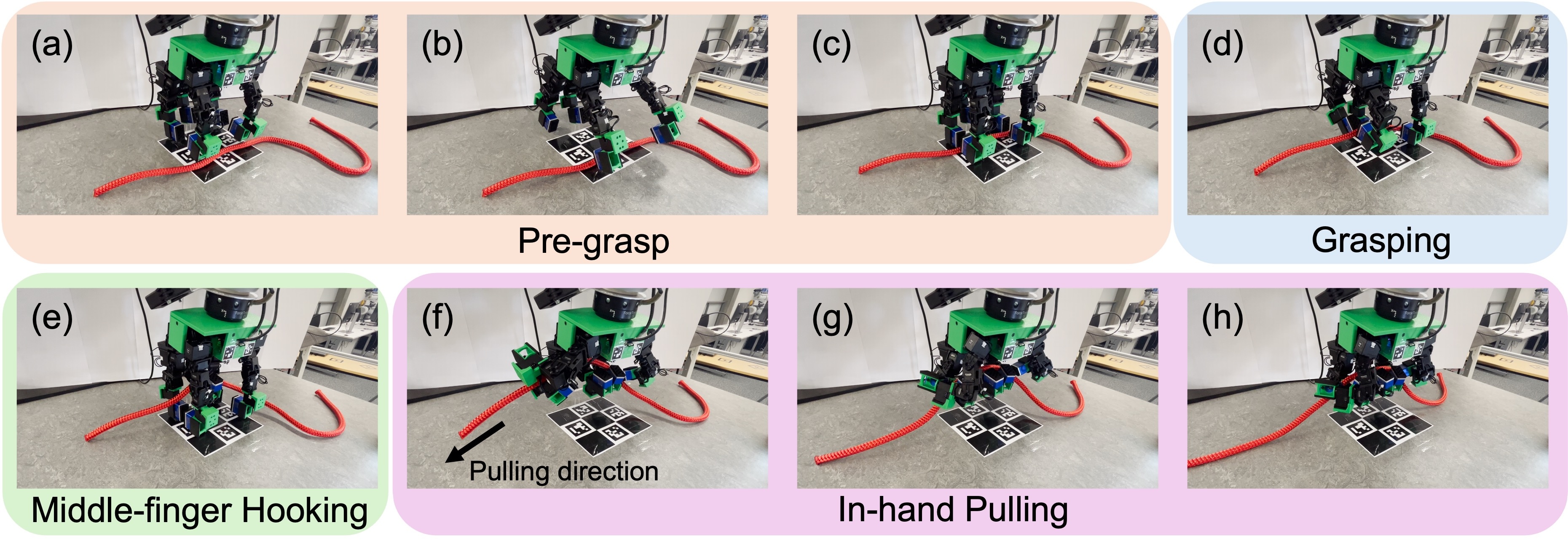}
\captionof{figure}{
Our designed hand pulling the cable from right to left. (a)-(c):  The hand performs a pre-grasp motion to position the cable correctly. (d): The left-side thumb and index finger grasp the cable. (e): The middle finger bends to hook the grasped cable. (f)-(h): The left-side thumb and index finger drag the cable to the left, and the right-side thumb and index finger hold the cable to prevent it from sliding back.
\label{fig:intro_demo}}
}]

\begin{abstract}
Humans use their hands to dexterously manipulate cables to perform various tasks, like grasping cables, moving cables in hand without dropping them, bending the cable into a U shape for hooking and so on.
Existing research that addressed cable manipulation relied on two-fingered grippers, which make it difficult to perform similar cable manipulation tasks that humans perform.
This is due to the limited dexterity of a two-fingered gripper, which can only grasp and release a cable without additional manipulability.
Thus, we need a multi-fingered hand, which is much more dexterous than a two-fingered gripper.
However, unlike dexterous manipulation of rigid objects, the development of dexterous cable manipulation skills in robotics remains underexplored due to the unique challenges posed by a cable's deformability and inherent uncertainty.
In addition, using a dexterous hand introduces specific difficulties in tasks, such as cable grasping, pulling, and in-hand bending, for which no dedicated task definitions, benchmarks, or evaluation metrics exist.
Furthermore, we observed that most existing dexterous hands are designed with structures identical to humans', typically featuring only one thumb, which often limits their effectiveness during dexterous cable manipulation.
Lastly, existing non-task-specific methods did not have enough generalization ability to solve these cable manipulation tasks or are unsuitable due to the designed hardware.
We address these three challenges in real-world dexterous cable manipulation in the following steps: 
(1) We first defined and organized a set of dexterous cable manipulation tasks into a comprehensive taxonomy, covering most short-horizon action primitives and long-horizon tasks for one-handed cable manipulation.
This taxonomy revealed that coordination between the thumb and the index finger is critical for cable manipulation, which decomposes long-horizon tasks into simpler primitives. 
(2) We designed a novel five-fingered hand with 25 degrees of freedom (DoF), featuring two symmetric thumb-index configurations and a rotatable joint on each fingertip, which enables dexterous cable manipulation. 
(3) We developed a demonstration collection pipeline for this non-anthropomorphic hand, which is difficult to operate by previous motion capture methods.
Given only one demonstration on one specific cable for each manipulation, among 8 primitives, our method achieved 88\% success rate of demonstration replaying on three cables of the same material but various diameters and over 75\% success rate on three cables of very different materials, stiffness, and diameters. 
Based on collected primitive demonstrations, we developed finite state machines (FSM) that enabled the robotic hand to execute complex long-horizon tasks without requiring prior demonstrations of the complete trajectories.
Among four long-horizon and complicated manipulations, we achieved a 64\% success rate of demonstration replaying on the cables of the same materials and various diameters.
Our robotic hand achieved performance comparable to human baseline dexterity in both primitive actions and long-horizon tasks.
Through human-guided transitions between primitives, the robotic hand demonstrates robust long-horizon manipulation capabilities even under diverse external disturbances.
\end{abstract}


\section{Introduction}
The manipulation of cables, deformable linear objects in a formal description, is a fundamental task across diverse settings - from precise movements in surgical procedures to everyday handling in offices and homes, as well as critical applications in manufacturing and various industries \cite{zhu2022challenges,sanchez2018robotic, yin2021modeling}.
Previous works demonstrated that cables can be manipulated in versatile but specific ways with one or two grippers mounted on robot arms, e.g., cable insertion \cite{yu2023precise, wang2015online, zhou2020practical}, cable sliding and following \cite{she2021cable, yu2024hand}, cable knotting \cite{moll2006path, kudoh2015air}, cable untangling \cite{viswanath2022autonomously}, cable waving \cite{chi2022iterative, wang2024self}, cable routing \cite{jin2022robotic, luo2024multi}, cable coiling and wrapping for packing and storing \cite{ma2023robotic, ma2022action}, cable shape control and planning in 2D and 3D\cite{yan2021learning, yu2022shape, lv2022dynamic} and so on.
However, a significant gap still exists between robotic cable manipulation and human-level dexterity.
We demonstrate the gap by considering a manipulation case: a human wants to insert a USB connector into a computer port, and they first grasp the middle part of USB cable, slide the cable to reach the USB port connector, adjusts the connector's position and orientation, and finish insertion.
In the above-mentioned task, humans use their dexterous hands to perform versatile manipulation skills, allowing them to handle cables in numerous ways without the need for specialized tools or being limited to particular tasks.
This versatility highlights the limitations of current robotic approaches, e.g. using a robot gripper, which lack the adaptability and dexterity inherent in human hands and cannot perform either in-hand manipulation or long-horizon manipulations.
Although some grippers with task-specific modifications can perform a certain dexterous cable manipulation, e.g. a gripper with tactile sensors for in-hand cable following \cite{she2021cable, yu2024hand}, such modifications highly depend on the hardware designs for specific tasks without generalization like human hands.  

One straightforward way to begin to bridge the gap between robot and human manipulation of cables is to analyze and understand how humans manipulate cables. 
Investigating robotic successes on dexterous manipulation on rigid objects demonstrates some commonalities:  multi-fingered hands were mainly used with high dexterity on tasks that were already well-defined in studies on human manipulation, e.g. in-hand object reorientation \cite{chen2023visual, chen2022system, handa2022dextreme}, multiple object grasping \cite{yao2023exploiting}, and so on. To explore how humans manipulate objects, previous researchers developed taxonomies of rigid object grasping and manipulation \cite{cutkosky1989grasp, feix2015grasp, krebs2022bimanual, arapi2021understanding}.
These taxonomies can help segment, classify, and label different types of primitives or grasping poses among datasets from humans and provide a better understanding and analysis of human behaviors, for the benefit of enlightened robotics research. 
By manually defining primitives, task and motion planning can be proposed to address long-horizon manipulation of rigid objects with low level primitives and high level planners \cite{garrett2021integrated, mao2023learning, mao2024dexskills, edmonds2019tale, triantafyllidis2023hybrid}. 

Inspired by this rigid object research, for \textbf{d}exterous \textbf{c}able \textbf{m}anipulation (DCM) we propose a \textbf{cable}'s \textbf{dex}terous manipulation tax\textbf{onomy} (\textit{Cable Dexonomy}, see Section \ref{sec:taxonomy}.).

Built on the taxonomy for DCM, the second goal is to develop an improved end-effector. 
Human hands can perform dexterous cable manipulation in many ways without needing specific end-effectors or limitations to particular tasks. 
As for robotic cable manipulation, there is much previous research into cable manipulation \cite{yu2023precise, wang2015online, zhou2020practical,she2021cable,moll2006path, kudoh2015air, yu2024hand,viswanath2022autonomously,chi2022iterative, wang2024self,jin2022robotic,ma2023robotic, ma2022action,yan2021learning, yu2022shape, lv2022dynamic}, but they all used either one or two two-fingered grippers, and most grippers grasp cables firmly without relative motions between the cable and the gripper unless released, except \cite{she2021cable, yu2024hand}, which used tactile sensors that allowed sliding between the cable and the gripper. 
Using dexterous hands to manipulate cables is still under-explored, and one notable reason is both the dexterous hand and the cable have exceptionally high degree-of-freedoms (DoFs), which make the control extremely difficult.

The example in Figure \ref{fig:intro_demo} shows that using a multi-fingered hand to manipulate the cable has many benefits. 
The hand manipulates the cable in a dexterous way by first pre-grasping the cable into a more convenient position, grasping it, hooking it with the middle-finger, and performing in-hand pulling to move the cable along the hand. 
If a two-fingered gripper were used instead of the hand, it would be necessary to move the whole arm to perform pre-grasp and grasping and it would also need support from the table to perform the cable sliding for releasing and re-grasping. 
These benefits also happen in different scenarios with similar behaviors. 

One notable thing about the hand shown in Figure \ref{fig:intro_demo} is the hand has two new features compared to anthropomorphic hands: two symmetrically placed thumbs and rotatable fingertips. 
These improvements were identified as a consequence of the taxonomy study.
This will be further discussed in Section \ref{sec:hardware}.

Given the \textit{Cable Dexonomy} and the hand designed for dexterous cable manipulation, the third problem arises: how to control the new hand to enable it to perform various DCM tasks. 
As far as we can tell, all previously published solutions are task-specific without having (or at least demonstrating) the potential to generalize to different tasks. 
Considering the enormous types of cables and associated dexterous manipulations, proposing new solutions for each new manipulation task is time-consuming and unrealistic.
For a general-propose solution for various manipulations, either optimization-based methods \cite{mordatch2012contact} or learning-based methods have demonstrated their potential \cite{petrenko2023dexpbt, nagabandi2020deep}. However, the former requires a careful construction of optimization equalities, and the latter one needs specific reward engineering and meanwhile sim-to-real transferring is another challenge \cite{andrychowicz2020learning, handa2022dextreme}. Recently, imitation learning-based methods have become very popular for robotic manipulation with single or dual arms, e.g. Action Chunking Transformer \cite{zhao2023learning, zhao2024aloha}, Diffusion Policy \cite{chi2023diffusion}, with a multi-fingered hand and Dexterous Imitation Made Easy \cite{arunachalam2023dexterous}, See to Touch \cite{guzey2024see}, on an anthropomorphic hand. 
However, these methods, requiring motion capture or teleoperation, are almost impossible to use for dexterous cable manipulation, which is a multi-contact problem and requires intensive haptic feedback. 
So far, we have not seen any method that can collect in-hand demonstrations of dexterous manipulation of cables with an anthropomorphic hand, let alone with a hand whose dexterity is beyond the usual anthropomorphic design.

Therefore, to address the control problem by using imitation learning, we proposed a special data collection pipeline, together with a finite state machine that can execute primitives to perform a long-horizon manipulation. (See Section \ref{sec:policy}.)

We do not propose or implement an autonomous manipulation system in this paper. 
The goal of the experiments is to show that: 1. The proposed demonstration collection pipeline is well established and capable of replaying primitive and long-horizon actions on different cables with different configurations. 
2. A good cable taxonomy could lead to a good performance of long-horizon manipulations with clearer strategies and sub-steps.
3. A good hardware design that can perform these dexterous manipulation with a high success rate.
4. The potential of collecting a large scale of demonstration dataset to train a learning-based manipulation policy.

To conclude, this paper makes three contributions:

\begin{itemize}
    \item A proposal for a dexterous cable manipulation taxonomy (\textit{Cable Dexonomy} - Section \ref{sec:taxonomy}).
    \item An improved multi-fingered hand with one additional thumb and rotatable fingertips in symmetric structure inspired by \textit{Cable Dexonomy}, which is capable of performing various dexterous cable manipulation tasks. (Section \ref{sec:hardware})
    \item An effective demonstration collection pipeline using the designed hand to identify and implement manipulation primitives and replaying long horizon tasks with finite state machines according to the \textit{Cable Dexonomy} with unseen cables. (Section \ref{sec:policy}).
\end{itemize}

\section{Research Background}
\label{sec:background}

\textbf{Dexterous manipulation} refers to turning and shifting of objects to change their positions and orientation in the hand through the motion of the palm and fingers given a reference configuration \cite{mason2018toward, bicchi2000hands}. Compared to gripper-based manipulation, dexterous manipulation usually involves a multi-fingered hand to perform in-hand manipulation, including dexterous grasping \cite{xu2023unidexgrasp}, in-hand object reorientation \cite{chen2023visual, chen2022system, andrychowicz2020learning}, in-hand finger motion planning \cite{gao2024enhancing}, grasped object shape estimation \cite{khadivar2023online}, tossing \cite{huang2023dynamic}, catching \cite{kim2014catching}, multiple object grasping \cite{yao2023exploiting}, solving Rubik's cubes \cite{akkaya2019solving}, rotating Baoding balls \cite{nagabandi2020deep} and so on. Another way of performing dexterous manipulation is called extrinsic dexterity \cite{dafle2014extrinsic}. With the help of gravity or extrinsic contacts with environment, like tables or walls, end-effectors can perform more dexterous manipulation including object flipping \cite{sun2020learning, hogan2020tactile}. In this work, we focus on dexterous manipulation with a multi-fingered hand, but we still discussed how contacts with environment benefit in-hand dexterous manipulation.

\textbf{Manipulation taxonomies} have been proposed to classify different types of human grasping and manipulation, due to the many and complex manipulations possible with different objects and with different goals. Cutkosky et al. first developed a taxonomy for power and precision grasps based on the concept of virtual fingers \cite{cutkosky1989grasp}. Feix et al. extended human grasping taxonomies into The \textit{GRASP} Taxonomy with 33 different grasp types \cite{feix2015grasp}.
Beyond human grasping, there are several works on taxonomies for  labeling and segmentation of videos of human object manipulation
\cite{krebs2022bimanual, arapi2021understanding, bullock2011classifying, bullock2012hand}. 
These manipulation taxonomies have two advantages: 1. a robotic manipulation benchmark, and 2. introducing long-term manipulation by composition of short-term primitives.

\textbf{Cable manipulation} has been explored in various way, including cable insertion \cite{yu2023precise, wang2015online, zhou2020practical}, cable following and sliding \cite{she2021cable, yu2024hand}, cable knotting \cite{moll2006path, kudoh2015air}, cable untangling \cite{viswanath2022autonomously}, cable waiving \cite{chi2022iterative, wang2024self}, cable routing \cite{jin2022robotic, luo2024multi}, cable coiling and wrapping for packing and storing \cite{ma2023robotic, ma2022action}, cable shape control in 2D and 3D\cite{yan2021learning, yu2022shape, lv2022dynamic} and so on. 
For these tasks, the most frequently used end-effectors are parallel jaw grippers or similar end-effectors that can firmly grasp the cables without relative sliding. 
However, fixed grasps of cables prevent the robot from performing DCM due to the restricted action space,  
She et al. \cite{she2021cable} used a gripper with tactile sensors to do cable following, allowing the gripper to reach one end-tip of the cable by sliding along it. 
Yu et al. \cite{yu2024hand} performed a similar manipulation using the thumb and a single index finger on a multi-fingered hand equipped with tactile sensors on the fingertips, but did not further explore the use of all fingers or other tasks.
For both methods, sliding the gripper along a cable, rather than pulling the cable through the hand, requires two strong assumptions: one end of the cable is fixed, and the hand is moved by the robot arm rather than staying static to perform in-hand manipulation.

In this paper, we show that using a 5-fingered hand without tactile sensors can perform not only one hand cable pulling, but also several other in-hand cable manipulation tasks.

\section{Taxonomy of Dexterous Cable Manipulation}
\label{sec:taxonomy}

\begin{figure*}
    \centering
    \includegraphics[width=0.97\linewidth]{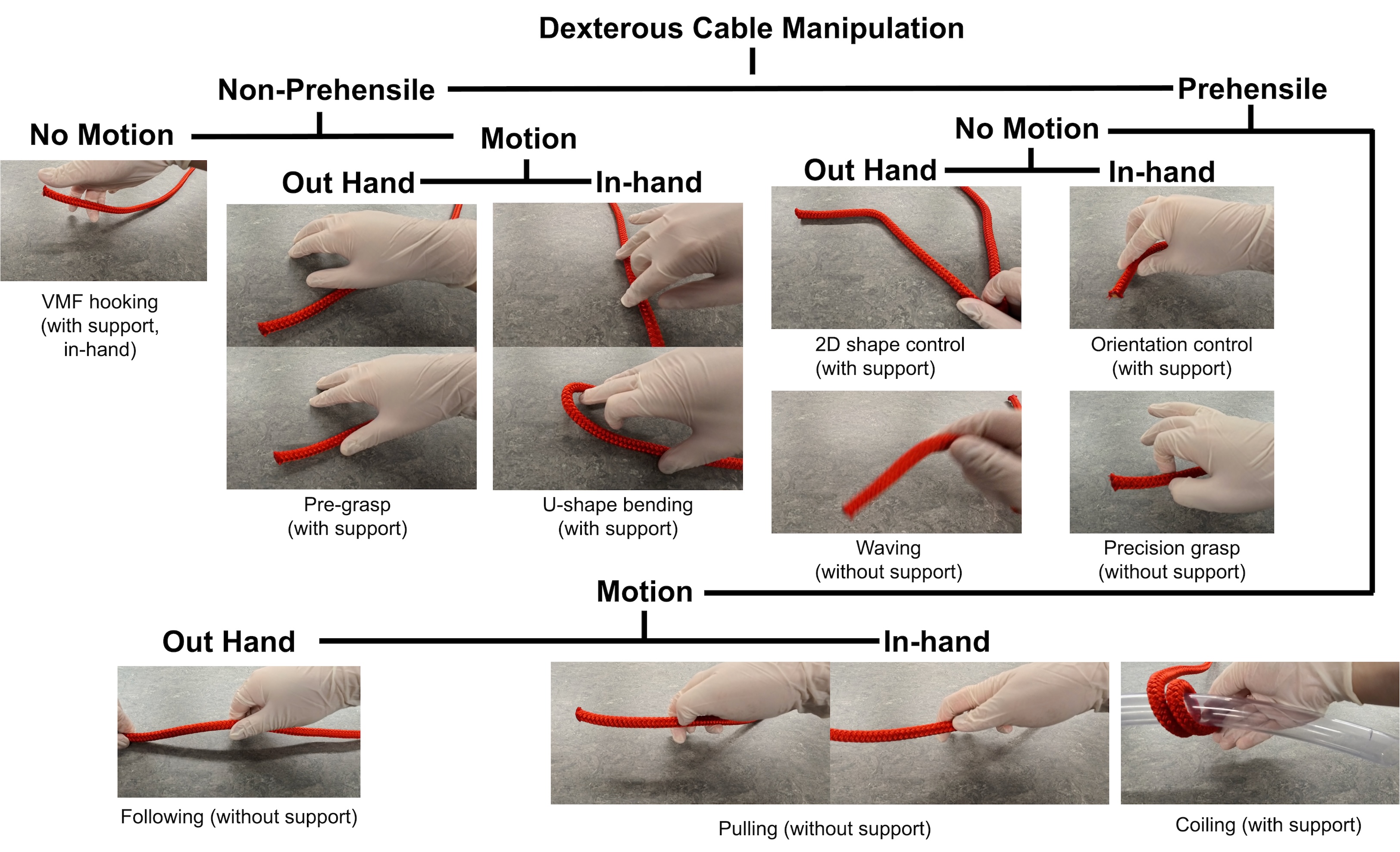}
    \caption{Human DCM with \textbf{Prehensile} (or not), \textbf{Motion} between fingers and the cable (or not), \textbf{In-hand} or out-of-hand, and \textbf{Support} from the external contact (or not).}
    \label{fig:tax_tree}
\end{figure*}

\begin{figure*}
    \centering
    \includegraphics[width=0.97\linewidth]{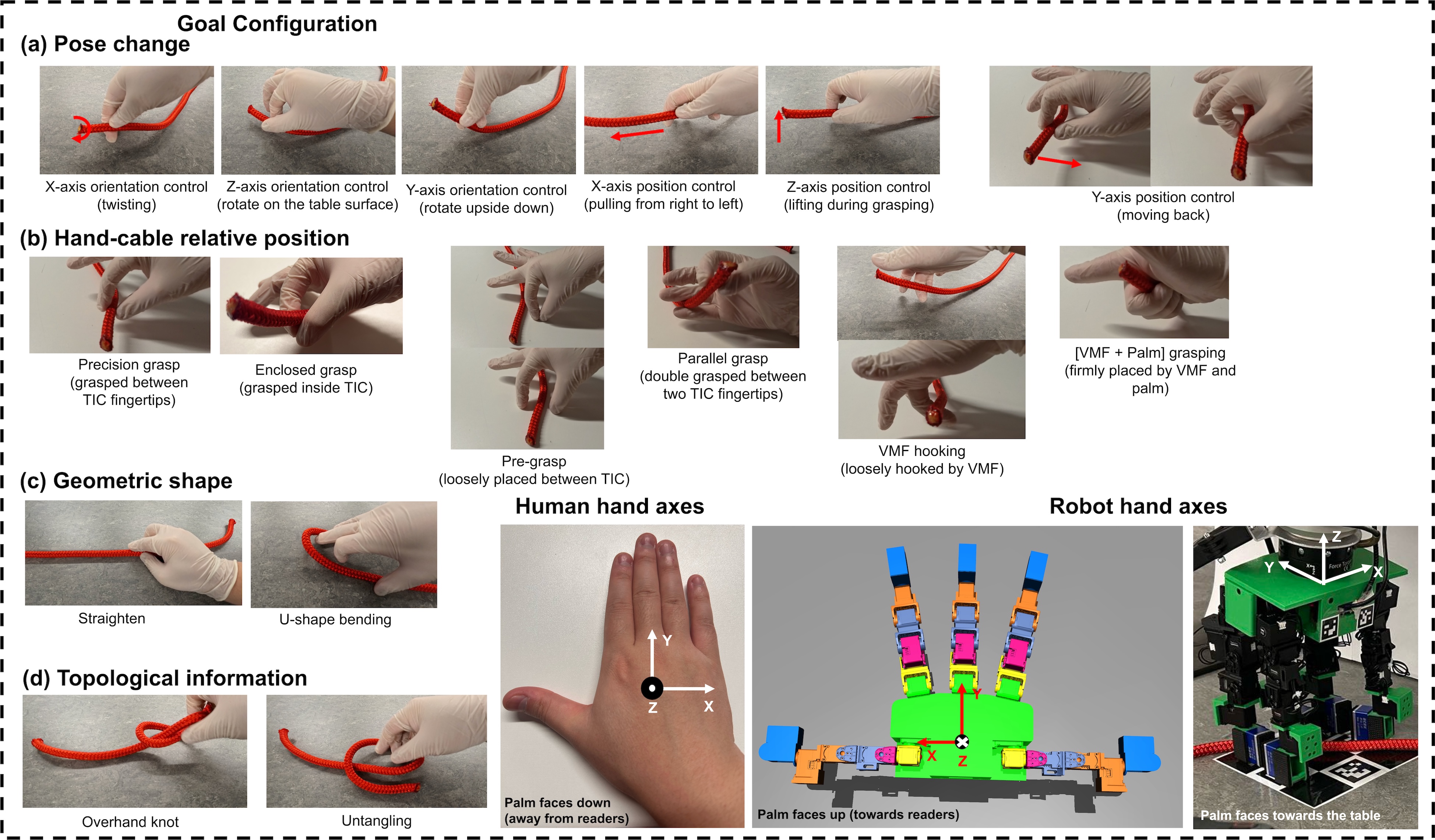}
    \caption{Four types of goal configurations. The coordinate system shown at the lower right is the same for the human hand, our rendered robot hand, and our real-world robot hand.
    }
    \label{fig:tax_goal_config}
\end{figure*}

This section introduces a proposed \textit{Cable Dexonomy} (Dexterous cable manipulation Taxonomy) from the human perspective.

\begin{table*}[h]
\centering
\caption{Example tasks classified according to 6 criteria. ? for unknown solutions because knotting or untangling normally requires two hands which are much more sophisticated than the case of using one hand. TIC is the combination of the thumb and the index finger, VMF is the virtual middle finger. $^*$ indicates that the cited research used a moving end-effector rather than a fixed-base. $^\dagger$ means that 3D shape control is not designed for single hand manipulation unless one end is fixed.}
\begin{tabular}{lllllll}
\hline
Example Tasks        & Prehensile                  & Motion                      & In-hand                     & Supporting                  & Fingers         & Goal Config       \\ \hline
Pre-grasp             & \XSolidBrush & \Checkmark   & \Checkmark   & \Checkmark   & TIC   & hand-cable        \\
Precision grasp      & \Checkmark   & \XSolidBrush & \Checkmark   & \XSolidBrush & TIC             & hand-cable / Z-position  \\
Parallel grasp       & \Checkmark   & \XSolidBrush & \Checkmark   & \XSolidBrush & TIC + [VMF+Palm]         & hand-cable / Z-position  \\
Precision to power   & \Checkmark   & \Checkmark   & \Checkmark   & \XSolidBrush & TIC + VMF       & hand-cable         \\
VMF hooking          & \XSolidBrush & \XSolidBrush   & \Checkmark   & \XSolidBrush &    TIC+ VMF             & hand-cable        \\
Pulling \cite{zhaole2023dexdlo}   & \Checkmark   & \Checkmark   & \Checkmark   & \XSolidBrush & TIC + [VMF+Palm]   & X-position  \\
Following \cite{she2021cable}$^*$, \cite{yu2024hand}$^*$          & \Checkmark   & \Checkmark   & \XSolidBrush & \XSolidBrush & TIC             & X-position  \\
Position control \cite{zhaole2023dexdlo}    & \Checkmark   & \XSolidBrush & \Checkmark   & either                        & TIC             & position    \\
Orientation control  & \Checkmark   & \XSolidBrush & \Checkmark   & either                        & TIC / TIC + VMF & orientation \\
2D shape control \cite{yan2020self}$^*$         & \Checkmark   & \XSolidBrush   & \Checkmark   & \Checkmark   & TIC         & shape             \\
3D shape control$^\dagger$ \cite{liu2023robotic}$^*$         & \Checkmark   & \XSolidBrush   & \Checkmark   & \XSolidBrush   & 2 * TIC         & shape             \\
Straighten          & \Checkmark   & \Checkmark   & \Checkmark   & \Checkmark   & TIC + [VMF+Palm]         & shape             \\
U-shape bending      & \XSolidBrush & \Checkmark   & \Checkmark   & \Checkmark   & TIC + VMF   & shape             \\
Overhand knotting    & \Checkmark   & \Checkmark   & \Checkmark   & either                        & ?               & topology              \\
Waving \cite{wang2024self}$^*$, \cite{chi2022iterative}$^*$, \cite{zhang2021robots}$^*$             & \Checkmark   & \XSolidBrush & \XSolidBrush & \XSolidBrush & TIC             & position    \\
Direction flipping   & \Checkmark   & \Checkmark   & \Checkmark   &  \XSolidBrush                           & TIC + VMF       & orientation \\
Coiling  \cite{ma2023robotic}$^*$            & \Checkmark   & \Checkmark   & \Checkmark   & \Checkmark   & TIC + VMF       & topology              \\
In-hand peg-in-hole  & \Checkmark   & \Checkmark   & \Checkmark   & \XSolidBrush & 2 * TIC         & pose    \\
In-hand pose control & \Checkmark   & \XSolidBrush   & \Checkmark   & \XSolidBrush & 2 * TIC         & pose \\
Routing by insertion \cite{luo2024multi}$^*$ & \Checkmark   & \XSolidBrush   & \Checkmark   & \XSolidBrush & 2 * TIC         & pose    \\
Fasten (a safety belt)           & \Checkmark   & \Checkmark   & \Checkmark   & \Checkmark   & TIC + [VMF+Palm]         & pose    \\
Untangling  \cite{viswanath2022autonomously}$^*$         & \Checkmark   & \Checkmark   & \Checkmark   & \Checkmark   & ?               & topology              \\ \hline
\end{tabular}
\label{table:tax}
\end{table*}

\subsection{Cable Dexonomy Tree}

\begin{figure}
    \centering
    \includegraphics[width=0.97\linewidth]{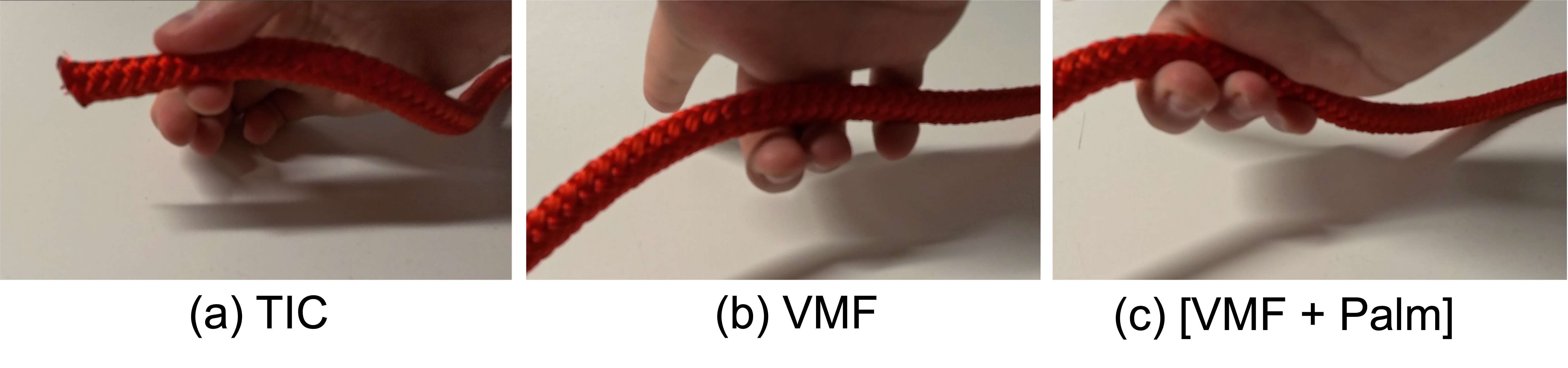}
    \caption{Three types of combinations of fingers. (a) TIC: the thumb and the index finger combo. (b) VMF: virtual middle fingers can be a combination of the middle finger, the ring finger, and the little finger. (c) [VMF+Palm]: using VMF and the palm for extra grasping, can be replaced by another TIC if exists.}
    \label{fig:tax_tic_vmf}
\end{figure}

Previously, a hand-centric manipulation taxonomy was proposed to classify different in-hand rigid object manipulation tasks \cite{bullock2012hand}, given 6 criteria used to define the task. 
There are 4 criteria shown in Figure \ref{fig:tax_tree}: prehensile or non-prehensile, motion or no motion, in-hand or out-of-hand, and with or without support. 
Figure \ref{fig:tax_tic_vmf}, shows that the fingers used during manipulation can be summarized into three types. 
Figure \ref{fig:tax_goal_config}, displays the different goal configurations that the cable may reach during dexterous manipulation.

Next, we describe how Bullock {\it et al}'s scheme \cite{bullock2012hand} is adapted for cables and dexterous manipulators. 
We propose several cable-specific extensions for the criteria mentioned above: 
1. \textbf{Prehensile} refers to having more than one contact point and having contact forces that can stabilize the cable without external forces like gravity or ground support \cite{bullock2012hand}. 
2. \textbf{Motion} indicates active movement exists between the hand and the cable. 
3. \textbf{In-hand} means the manipulation mostly happens inside the hand space between fingers and the cable, and its opposite, out-of-hand, refers to the movement of the whole arm or the wrist without considering the fingers' actions. 
4. \textbf{With support} means that the cable is supported by an external contact, such as lying on the table or the ground. 
5. \textbf{Used fingers} indicates which fingers and how many are used to perform the manipulation. 
Two additional terms are introduced: virtual middle finger (VMF) and thumb-index combination (TIC).
Virtual fingers (VF) were proposed by Cutkosky \cite{cutkosky1989grasp} and used in the \textit{GRASP} taxonomy \cite{feix2015grasp}, which refers to several fingers working together as a functional unit. More specifically, a VMF includes at least one middle finger possibly with the ring finger and the little finger, shown in Figure \ref{fig:tax_tic_vmf}.
The thumb-index combination (TIC) refers to the most frequently used two fingers among many tasks, i.e. the thumb and the index finger. 
[VMF+Palm] indicates using the VMF and palm to grasp the cable without the ability to control the cable's pose (shown in Figure \ref{fig:tax_tic_vmf}), whereas one TIC can grasp the cable with the ability to control its pose. Different from \textit{The Grasp Taxonomy}, we do not assign the palm to a virtual finger.
6. \textbf{Goal configuration} describes the target state of the cable. 

The configuration descriptions of the 4 types are: 1) pose change of the local part, 2) overall geometric shapes, 3) hand-cable relative position, and 4) topological information.
Though a long thin cylinder's motion was discussed by Bullock et al. \cite{bullock2012hand}, cables are much more complicated than cylinders due to their relative smaller radius and possible shape deformation. 
`Pose change' refers to a local movement of the cable in orientation or position in three dimensions. 
`Overall geometric shapes' refers to the cable shape control problem, such as making a cable straight or into a U-shape. 
The `hand-cable relative position' refers to how the cable is placed with respect to the hand, e.g., the cable is placed between the thumb and the index finger for future grasping. 
`Topological information' refers to adjacency relations among intersections during knotting and untangling \cite{takamatsu2006representation, grannen2020untangling}. 
Figure \ref{fig:tax_goal_config} shows the criteria for these four types of goal configurations. 
Some primitives are complicated so they may have multiple goal configurations.
For example, the goal configuration of precision grasping usually contains hand-cable relative position and a Z-axis position control by lifting the cable.

\subsection{Common Manipulation Tasks}
Table \ref{table:tax} displays some common manipulation tasks and their categories according to the proposed \textit{Cable Dexonomy}, together with their corresponding research which used grippers or a multi-fingerd hand that can be moved by the robot arm. 
Because a jaw-parallel gripper grasps the cable firmly during manipulation, a human hand can also perform this with one TIC or [VMF + Palm]. Thus, our \textit{Cable Dexonomy} also covers some previous studies that used jaw-parallel grippers by clarifying the used fingers. Some examples are:
\begin{itemize}
    \item 2D shape control on the table: Yan et al. \cite{yan2020self} used a gripper to control the cable's 2D shape by picking and placing. This can be performed with a TIC using a precision grasp.
    \item 3D shape control in the air: Yu et al. \cite{yu2022shape} used dual grippers with two arms to control the cable's 3D shape. With two TICs or two [VMF + Palm] that can freely move in the air, we can still perform 3D shape control.
    \item Cable pulling: Two grippers can grasp and release the cable as an alternative way to move it to the target position, and meanwhile two TICs can perform the same behavior.
    \item End-tip orientation control: One gripper/TIC that is attached to a robot arm can freely control the orientation of the cable's end-tip. 
    However, if the hand base is fixed, we use a TIC to grasp the cable, and VMF to pivot it by applying a force on the other side. This can be performed with a gripper loosely grasping the cable (This can be achieved by a gripper with tactile sensors \cite{she2021cable}.) and another additional finger to pivot the other side of the cable to rotate its end-tip.
\end{itemize}

These cases indicate that cable manipulation with one or multiple grippers mounted on a robot arm can be performed using a multi-fingered hand such as the one presented here.

\subsection{Composition of Long-Horizon Tasks}
\begin{figure}
    \centering
    \includegraphics[width=0.97\linewidth]{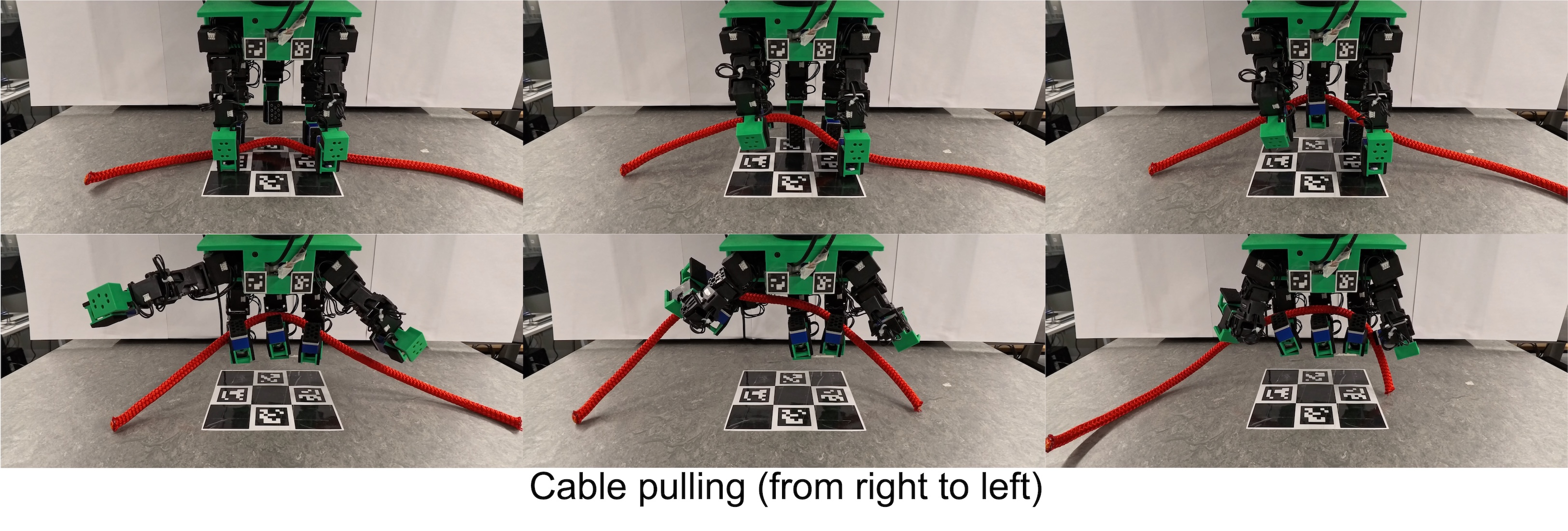}
    \caption{Demonstrations of cable pulling. See the appendix for other long-horizon manipulations.}
    \label{fig:tax_long_horizon_task}
\end{figure}

Long-horizon tasks are those that require a sequence of short-term actions (primitives). 
They are mostly complex and require a high-level planner to execute primitives in the correct order  \cite{garrett2021integrated, bullock2012hand}. 
How to decompose a long-horizon task is crucial to solve the task. 
The primitive actions of the \textit{Cable Dexonomy} allow the decomposition of many long-horizon tasks into a sequence of several primitives. 
One long-horizon task example is Cable pulling, as in Figure \ref{fig:tax_long_horizon_task}.
The hand performs pre-grasp, grasping, middle-finger hooking, and cable pulling repetitively until reaching the end-tip.

\section{Hardware Design}
\label{sec:hardware}
\begin{figure}
    \centering
    \includegraphics[width=0.97\linewidth]{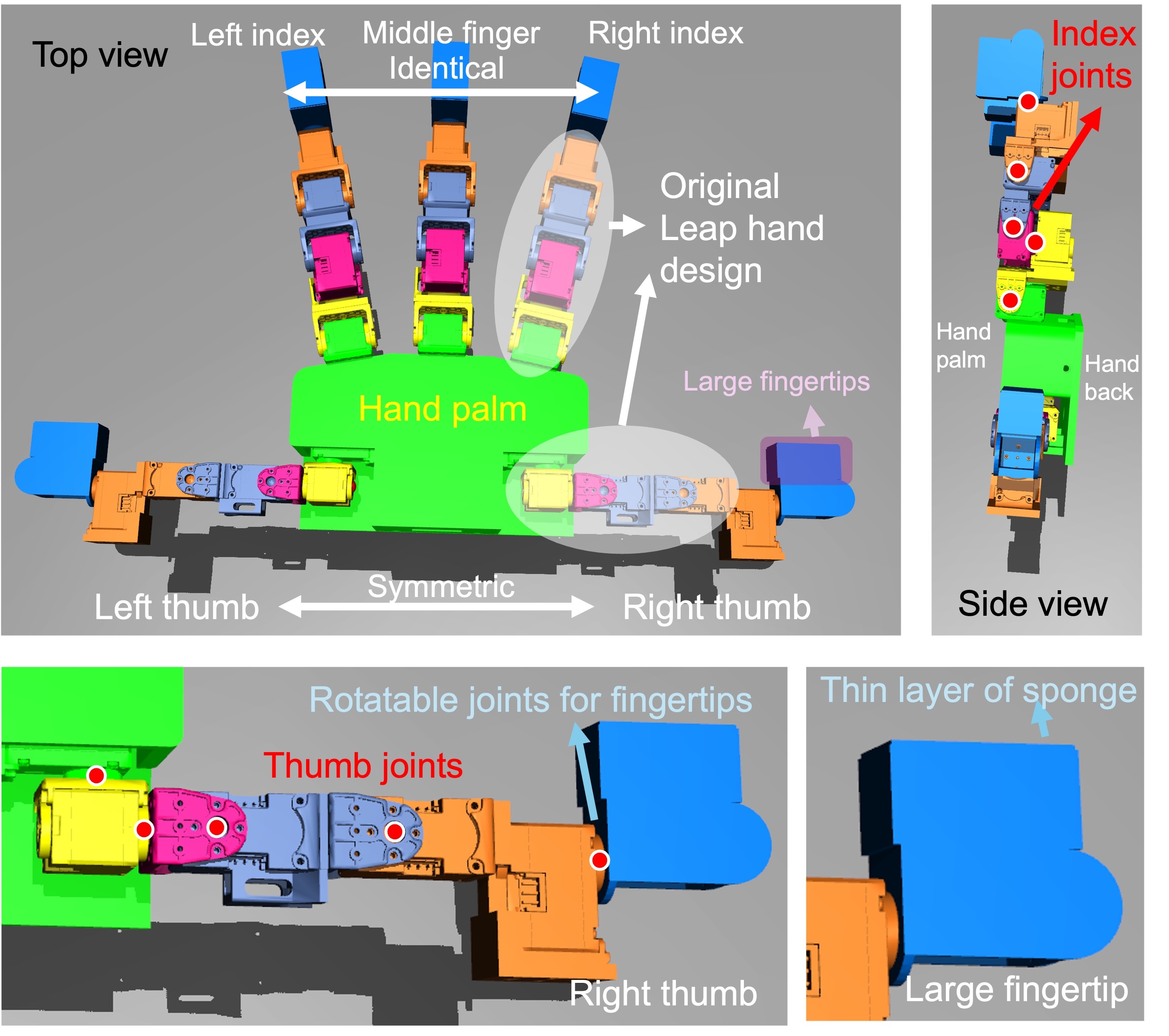}
    \caption{A top view from the hand palm side and a side view from the right thumb side of the 2-thumbed hand rendered in MuJoCo. The red dot indicates the joint positions on each identical finger. The last blue link is the rotatable fingertip. The white areas are the finger designs from the Leap Hand. The two thumbs have symmetric structures and the other three fingers have identical structures. The hand back is mounted to the last joint of the robot arm. All fingertips are the same size with a thin layer of sponge on each.}
    \label{fig:hardware_robothand}
\end{figure}

Many other hands are not open-source or available for purchase yet, such as the  Tesla Optimus's hands. 
These hands have the closest similarity to human hands: one thumb and a subset of the remaining four fingers, and are named \textit{Anthropomorphic} hands. 
There are also other multi-fingered hands that are not human-like. D'Claw hands \cite{ahn2020robel} have four identical index fingers, evenly distributed on a circle. Among these multi-fingered hands, the Leap hand and the D'Claw hand are often the best choices because they are cheap, open-sourced, and dexterous with high DoFs. However, the original D'Claw only has three joints on each finger, and identical fingers without any thumb make pincer grasping difficult, so we chose to use the Leap Hand as the basis for an extended hand design.

We use the same thumb and index fingers from the Leap Hand \cite{shaw2023leap}, but made several changes.
Although we still use \texttt{DYNAMIXEL XC330-M288-T} servo motors as the actuators, we used 25 motors instead of 16 motors which the Leap Hand used.
Two thumbs are designed in symmetric positions and the hand has five fingers in total.
Each finger has one additional rotatable fingertip, and therefore contains five actuators.

This section answers two questions: 1. Why is the original Leap hand (and other commonly used anthropomorphic hands) not a suitable design for dexterous cable manipulation (DCM)? 2. How can the Leap style of hands be improved for better cable manipulation?

\begin{figure}
    \centering
    \includegraphics[width=0.97\linewidth]{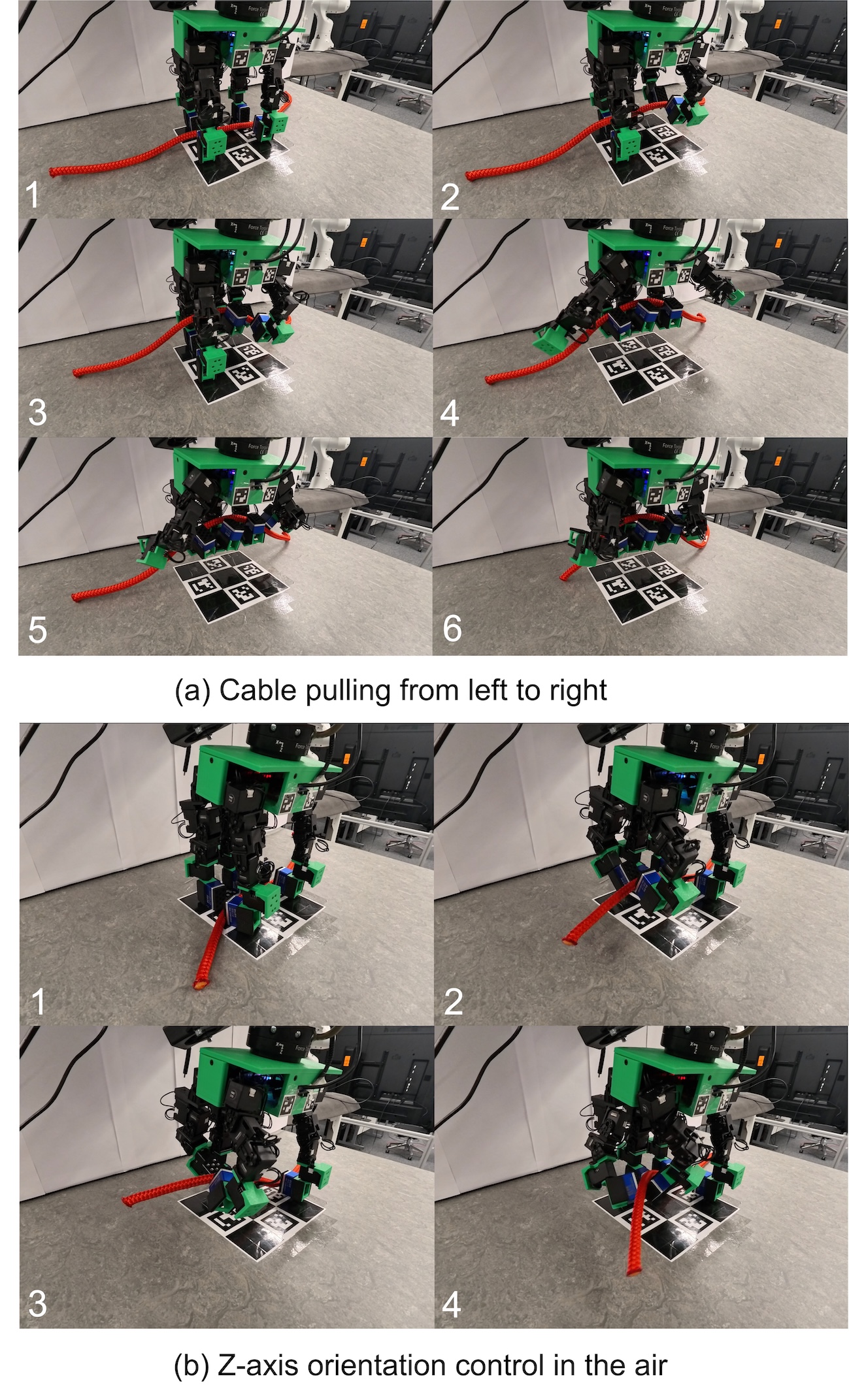}
    \caption{Two manipulations to display advantages of our designs, where each step is arranged in numerical order displayed in each image. (a) Cable pulling from left to right with the help of the additional thumb and the symmetric structure with the collected demonstration which is in a different pulling direction. (b) Z-axis orientation control in the air with the rotatable fingertips to avoid relative displacement between the cable and two fingertips which can cause cable slipping away.}
    \label{fig:hardware_two_advantages}
\end{figure}

\subsection{Dual-Thumb}

Inspired by the \textit{Cable Dexonomy} in Section \ref{sec:taxonomy}, we summarize the frequency of used fingers in different primitives: 
1. TIC is the most frequently used finger configuration, and almost all manipulations require TIC.
2. Most DCMs are symmetric about the Y-axis, e.g., the only difference between cable pulling from left to right and from right to left is the pulling direction. 
However, if an anthropomorphic hand that is asymmetric about the Y-axis is used, two different control policies are needed to perform the same tasks with different manipulation directions. Zhaole et. al explored this in the simulation \cite{zhaole2023dexdlo}. 
Another example is performing USB insertion, where the plug is on the other side of the TIC. 
Here, the hand needs to flip upside down to move the TIC near to the plug, because the little finger, which grasps the cable against the palm, does not have the ability to control the cable orientation.
For DCMs that are sensitive to the TIC's location, implementing two TICs in a symmetric layout is a good solution to keep the symmetric manner of DCM (see the design in Figure \ref{fig:hardware_robothand}). 
We also show a benefit that symmetric manipulations only require one demonstration for two opposite direction manipulations in Figure \ref{fig:hardware_two_advantages} a.

A notable work added an additional prosthetic thumb onto a human hand, which can affect object manipulability \cite{clode2024evaluating}, including holding a cup to do in-hand liquid pouring into the grasped cup. 
The multi-fingered design proposed here is the first one to confirm that adding an additional thumb can help extend robotic manipulability (as contrasted to the human manipulability explored in the previous research).

\subsection{Rotatable Fingertips}
When a human hand performs a pincer grasp on a cable, the skin of the fingertips will deform to cover the surface of the cable to make a larger contact area for more robust grasping. 
Thus, designing a better fingertip can increase the stability of cable grasping. We add an additional rotation joint to the fingertip to make such a grasping pose available in more configuration states.
When performing cable orientation control, the rotatable fingertips also help rotate the cable about the Z-axis without relative sliding between the cable and two fingertips, which cannot be performed by a cylinder-shaped fingertip.
Overall, one additional joint added to each hand can provide more manipulability.

Many tactile sensors have a relatively flat surface, like GelSight \cite{yuan2017gelsight}. 
Although we do not mount any tactile sensors, our rotatable fingertip design has the potential of mounting flat tactile sensors on it. E.g. NeuralFeel \cite{suresh2023neural} rotated their ring finger’s fingertip 90 degrees towards the in-hand operation space. With rotatable fingertips, such adjustments can be performed during manipulation in real-time.
In additional to Z-axis orientation control on the table, our hand performed this task in the air, shown in Figure \ref{fig:hardware_two_advantages} b.
When rotating the cable, if there is relative displacement between the fingertip and the cable, the cable will slowly slip downward and eventually drop from the TIC.
The additional rotatable joint on the fingertip avoids this and provides better manipulability in such a difficult task.

\section{Demonstration Collection Pipeline}
\label{sec:policy}
This section introduces the proposed demonstration data collection pipeline. There are three stages: 1. human short-term primitive demonstration data collection, 2. demonstration replay, and 3. building finite-state machines for long-horizon manipulation.

\subsection{Demonstration Collection}
\begin{figure}
    \centering
    \includegraphics[width=0.97\linewidth]{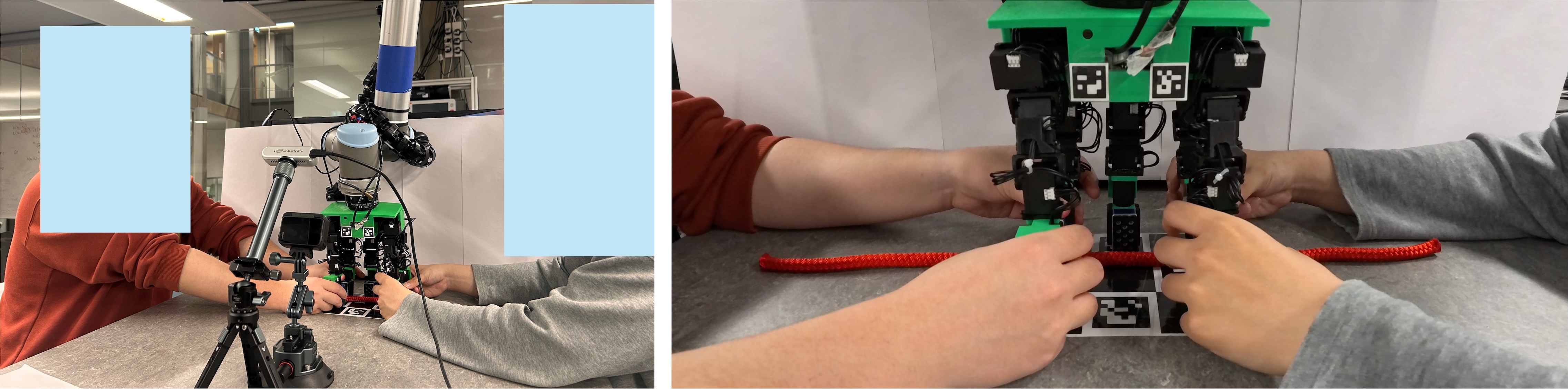}
    \caption{Two demonstrators drag the hand fingers to record the finger joint trajectories.}
    \label{fig:exp_human_demonstrator}
\end{figure}

Since the proposed hand has a different structure than the human hand and more degrees of freedom, common demonstration methods are not applicable, like motion capture with a mocap glove or a RGBD camera \cite{arunachalam2023dexterous, qin2022dexmv}. Instead, one or more people cooperate to gently drag the robot fingers in the same style as kinesthetic teaching of robot arms, shown in Figure \ref{fig:exp_human_demonstrator}.

The goal joint angle positions are set to be equal to the actual joint angles with a relatively low joint stiffness to make dragging easier and smoother. Once one or two fingers are temporarily fixed, we pause the collaborative mode and increase the joint stiffness to avoid unnecessary finger movements. During human demonstration data collections,  only  the joint angle trajectories are recorded. These trajectories will be used as actions in the demonstration replay later. 

The advantages of using dragging for demonstrations are: 
1. finer motor control recording of those dexterous manipulations than previous mocap-based methods, 
2. eliminates the need for mapping from the human finger joints to the robot hand joints, and most importantly 
3. it allows recording demonstrations from designs which are different from human hand structures. 
Two human demonstrators are needed to drag all five fingers, one to control one pair of TIC and the middle finger, and another to control the other TIC. 

\subsection{Demonstration Replay}
In this stage, the demonstration data is replayed according to the collected joint angle trajectories without any changes or human interventions to identify the successful demonstrations except for some compensation to the thumb and index finger joints  due to the low maximum torque of the hand motors.
PID position control was used to make the motors follow the joint trajectories from the demonstrations, with a control rate of 30 Hz.

This includes the first and the third joints (the order is determined from the proximal to the distal on each finger) of two index fingers and the second and the third joints of two thumbs.
We only recorded the first successful demonstration without evaluating its robustness in other scenarios or on other cables, and only the actual joint angles and cable key point positions were recorded. 
Each collected human demonstration can be replayed several times with different initializations of cable shapes and positions for more data. 
Although we do not apply learning-based methods here, the collected data can possibly be used to train the agent with an imitation learning algorithm, such as ACT \cite{zhao2023learning, zhao2024aloha} and Diffusion Policy \cite{chi2023diffusion}, or one-shot reusable learning strategies \cite{mao2023learning}.

The replay stage will show that the hand is capable of performing DCM and that a usable demonstration data collection is available (which may also be suitable for learning-based algorithms).

\subsection{Long-Horizon Manipulation}
\label{sec:demo:long_horizon}

Collecting demonstrations of long-horizon manipulation is difficult, since any failure that occurs  during the demonstration leads to total failure. 
Based on the \textit{Cable Dexonomy}, long-horizon DCM instances are decomposed into a sequence of basic primitives. To reduce the collection time  and increase the success rate of demonstration data collection,  only  short-term primitives were collected and by combining these allows long-horizon manipulations without prior full long-horizon demonstrations. 

Four long-horizon tasks are considered here, each with their own Finite State Machine (FSM): 1. Cable pulling, 2. Cable U-shape bending and grasping, 3. Cable direction flipping, and 4. Cable in-hand insertion. 
The FSM graph of cable pulling is shown in Figure \ref{fig:fsm_long_horizon_fsm}, and the rest FSM graphs and experiment figures can be found in the appendix and the video. 

With the help of the FSM, we demonstrate that the new hand is capable of performing long-horizon tasks. The demonstration data collection of long-horizon tasks is also available. 

\begin{figure}
    \centering
    \includegraphics[width=0.97\linewidth]{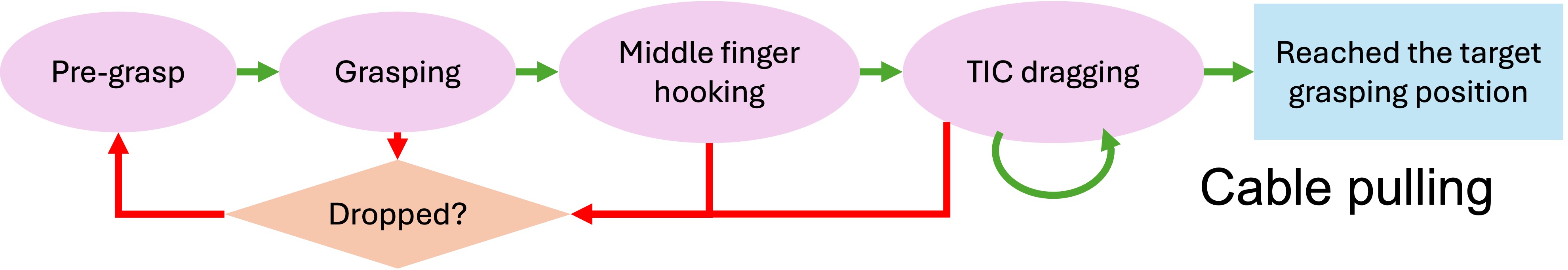}
    \caption{\textit{Cable Dexonomy} Finite State Machine for Cable Pulling.}
    \label{fig:fsm_long_horizon_fsm}
\end{figure}

\section{Experiments}
\label{sec:exp}
This section presents real-world experiments applying  different primitives to different cables.

\subsection{Experiment Setup}

The hand and the camera are placed as shown in Figure \ref{fig:exp_setup} a. Here are some details:

\begin{figure}
    \centering
    \includegraphics[width=0.98\linewidth]{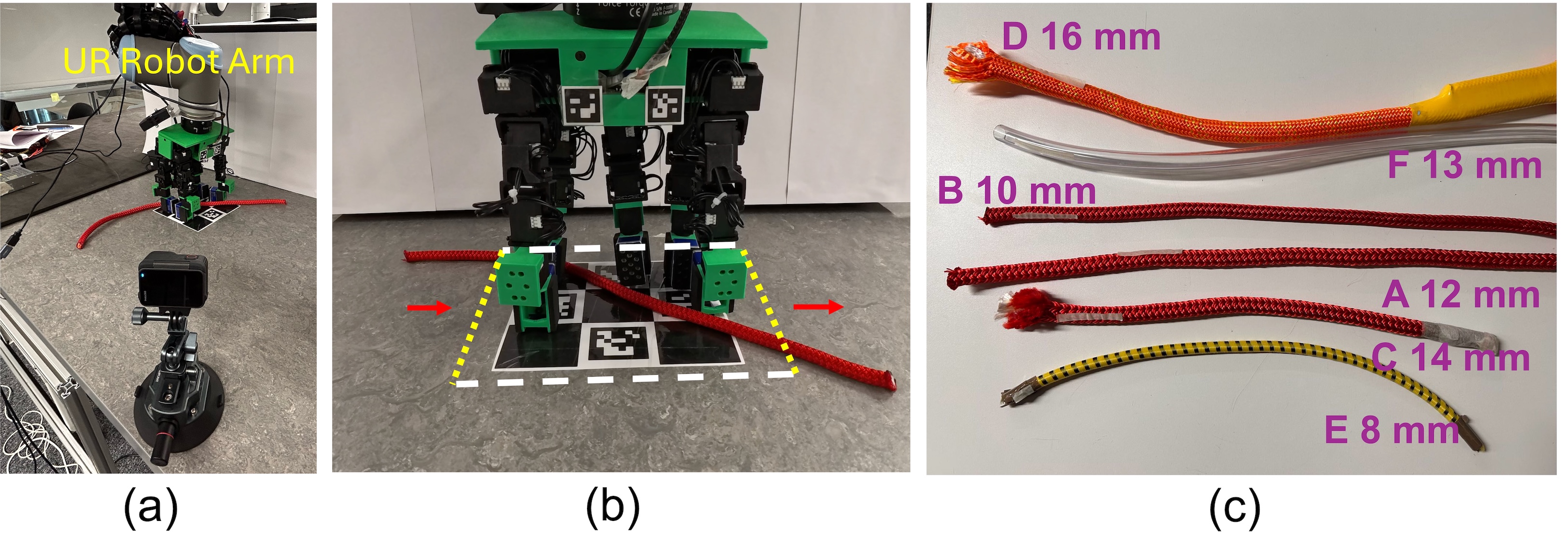}
    \caption{Experiment workspace setup. (a) The hand is mounted on the UR10 robot arm, which is kept static during the experiments with the palm facing down. (b) The cable is randomly initialized inside the white bounding box on the table, constrained to pass through the left and right side of the box, highlighted by the yellow dashed lines. (c) Different cables are used to evaluate the performance of each primitive. Only the red 12mm Cable A is used for demonstration data collection. Cable B and C have the same materials as Cable A but different diameters. Cable D, E, and F have very different materials and different diameters.}
    \label{fig:exp_setup}
\end{figure}

\textbf{Workspace.} The hand base is placed in a tilted pose to let the index finger and thumb finger exactly hang above the table without collision during manipulation. This can be seen in Figure \ref{fig:exp_setup} b. The cable was randomly placed in the in-hand space, also shown in Figure \ref{fig:exp_setup} b, and it is not always placed between the thumb and the index finger. 
The front and the back cable placement limits are determined according to the human demonstrations, to make sure the cable is not initialized too far from the hand, and the left and the right boundaries are set to 1 cm away from the left and the right thumb. 
Several different cables are used in the experiments, although only one is used for data collection. The evaluation cables are shown in Figure \ref{fig:exp_setup} (c) with their diameters in the figure.  Demonstration data was only collected using the 12mm red cable. 
The other two red cables with 10mm and 14mm diameters have the same material as the 12mm red cable. The 8mm yellow cable is an elastic cable for climbing, the 13mm transparent tube is made of plastic, stiff and hollow, and the 16mm thick cable has large stiffness.

\textbf{Demonstration Data Collection Instructions.} 
One demonstrator stands on the left side of the hand, and the other stands on the right. We use at most four hands to manipulate at most five fingers simultaneously. 
We replayed the demonstration with a similar initialization and the successful demonstrations are used for the later replay and performance evaluation. 
As mentioned previously, only  the first successful demonstration was recorded for use in the rest of the experiment. 
The recorded demonstration may perform poorly in other scenarios.

\textbf{Initialization.}
For the pre-grasp primitive, the cable was placed randomly according to the bounding box shown in Figure \ref{fig:exp_setup} b. 
The exact size and position of the bounding box is determined by the human demonstrators.
If Cable A is placed outside the bounding box, e.g. Cable A penetrates the white dashed line in Figure \ref{fig:exp_setup} b, the human demonstrators are unable to manipulate fingers to pre-grasp it.

For the other primitives, the cable was placed between two thumbs and two index fingers for simplicity, since the cable is always robustly pre-grasped into a suitable initialization position between the two thumbs and the two index fingers.
The difference between each trial is the initial position in the in-hand space.

\textbf{Evaluation metric.} 
Success rate $S$ is the evaluation metric, where
$S$ is the number of successful manipulations performed out of the total number of manipulation attempts. 
For most tasks, the grasping position is above the table and dropping the cable  from the hand is a failure.
If the cable cannot be manipulated to the goal configuration by the human hand within 10 seconds, it  counts as a human performance failure.
Similarly, a robot failure occurs if a success does not occur within a fixed number of robot hand manipulation steps. 
Each experiment has 5 different trials with slightly different cable position initializations.
 
To compare the robot hand's dexterity with humans, we need to first restrict human hands' dexterity because  if the dexterity and perception of the human hand are not restricted, the human hand will undoubtedly have an overwhelming advantage, and the experimental results compared with such baseline results will be meaningless. 
We propose a lower bound on human dexterity as our task baseline:
by wearing a thick skiing glove on the non-dominant hand, which is used less frequently than the dominant hand during daily life, the human operator receives much less tactile sensing and has a reduced hand dexterity.
The skiing glove has a similar size to the proposed robot hand, shown in Figure \ref{fig:exp_primitive_demo} a and b.
After taking a look at the cable and the hand, the human operator will close their eyes and manipulate the cable according to the given tasks to avoid the benefit from the visual feedback.

The approximate lower bound on the dexterity of the human hand through these restrictions gives some indications on how far the dexterity and manipulability of the proposed robot hand and algorithms are from human performance in dexterous cable manipulation tasks.

\subsection{Performance of Demonstration Replay}

The DCM demonstration replay performance was evaluated on 8 short-term primitives and four long-horizon tasks.
Here are the details of each primitive's successful goal configuration.

\textbf{1) Pre-grasp}: the cable is placed between the two thumbs and two index fingers (Figure \ref{fig:intro_demo} a-c).

\textbf{2) Precision grasp}: one TIC grasps the cable and lifts it more than 3cm higher than the original height without dropping. The precision grasp is also the Z-axis position control (Figure \ref{fig:intro_demo} d).

\begin{figure*}[h]
    \centering
    \includegraphics[width=0.97\linewidth]{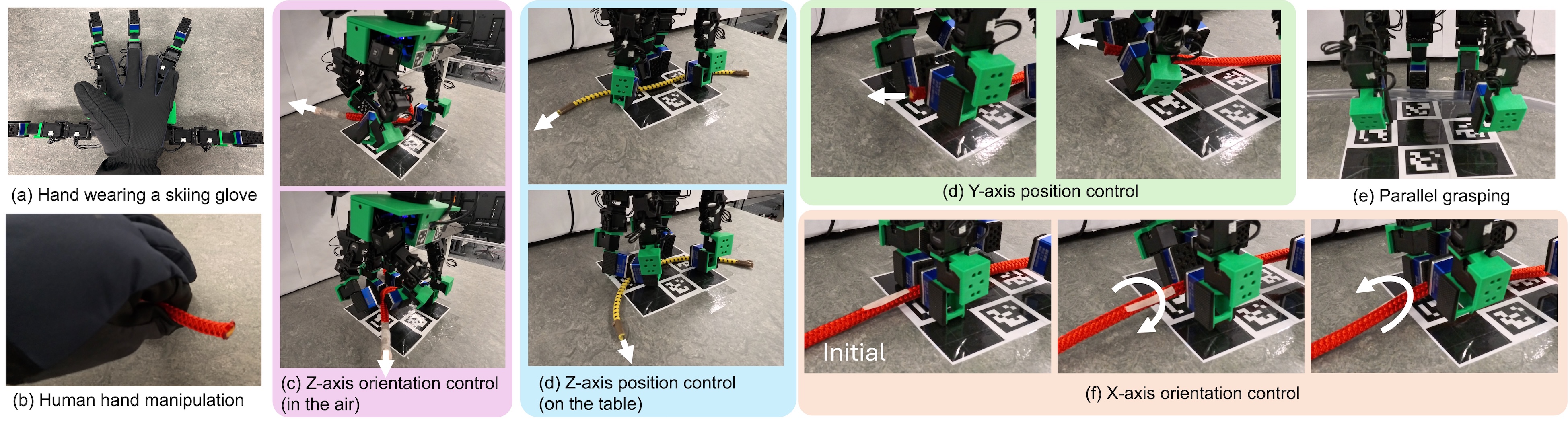}
    \caption{(a)-(b) Human non-dominant hand wearing a skiing glove as our baseline. (c)-(f) Short-term primitives of different cables performed by our hand, where the cable's end-tip pose is highlighted with the while arrows in each figure.}
    \label{fig:exp_primitive_demo}
\end{figure*}

\textbf{3) Parallel grasp}: the same goal configuration as a precision grasp, except two TICs need to grasp and lift the cable (Figure \ref{fig:exp_primitive_demo} e).

\textbf{4) VMF hooking}: the middle finger hooks the grasped cable to place it between the fingertip of the middle finger and the middle metacarpal link (Figure \ref{fig:intro_demo} e and Figure \ref{fig:tax_long_horizon_task}).

For the following position and orientation control primitives, we do not set a specific angle or distance because we can always stop the hand motion early to achieve the specific goal (see different goal configurations in Figure \ref{fig:tax_goal_config}). 

\textbf{5) X-axis orientation control}: the cable is twisted around the X-axis by more than 22.5 degrees in both directions (Figure \ref{fig:exp_primitive_demo} f). 

\textbf{6 \& 7) Z-axis orientation control}: the cable is rotated around Z-axis by more than 22.5 degrees in both directions. 
6) In the air means that the cable needs to be rotated while grasped in the air without dropping (Figure \ref{fig:exp_primitive_demo} c).
7) On the table indicates that the cable can be orientated with the support from the table without the need of being grasped, which is much easier than being rotated in the air (Figure \ref{fig:exp_primitive_demo} d).

\textbf{8) Y-axis position control}: the cable is moved in the Y-axis direction with a moving range of more than 4cm (Figure \ref{fig:exp_primitive_demo} d).

There are four long-horizon tasks:

\textbf{1) Cable pulling}: The cable can be pulled fully from the left to the right in the air without dropping. The cable pulling is also a form of X-axis position control.

\textbf{2) Direction flipping}: the cable's left end-tip which initially faces the left side can be turned in the air to face the right side.

\textbf{3) U-shape bending and grasping}: the cable can be first bent with the middle finger into a U-shape, and parallel grasped by two TICs.
The furthest part of the cable between the two TICs is at least 5cm from the hand center in the Y-axis direction.

\textbf{4) In-hand insertion}: the right TIC first grasps the 2.4cm diameter hollow tube. The left TIC grasps the cable and inserts it at least 2cm into the tube.

We did not add Y-axis orientation control because we could not generate a successful demonstration.

\begin{table*}[ht]
\centering
\caption{Success rate $S_{robot}/S_{human}$  of each primitive and long-horizon tasks. *Cable pulling by the human occurs in two directions, from the dominant hand side to the non-dominant hand side and the reverse direction, and so there are 10 trials in total. 
}
\label{table:exp_performance}
\begin{tabular}{lllllll}
\hline
        & 12mm                        & 10mm                        & 14mm                        & 16mm                        & 8mm                         & 13mm                               \\
                             & Red A                       & Red B                       & Red C                       & Orange D                    & Yellow E                    & Tube F                           \\
Short-term Primitives                             & \multicolumn{3}{c}{Easy cables}                                                         & \multicolumn{3}{c}{Hard cables}                                                                     \\ \hline
Pre-grasp                    & 1.0/1.0 & 1.0/1.0 & 1.0/1.0 & 1.0/1.0 & 1.0/1.0 & 1.0/1.0 \\ 
Precision grasp              & 1.0/1.0 & 1.0/1.0 & 1.0/1.0 & 1.0/1.0 & 0.8/0.0 & 1.0/1.0 \\ 
Parallel grasp               & 1.0/1.0 & 1.0/1.0 & 1.0/1.0 & 1.0/1.0 & 0.6/0.4 & 1.0/1.0 \\ 
VMF hooking                  & 1.0/1.0 & 1.0/0.6 & 1.0/0.6 & 1.0/1.0 & 1.0/0.0 & 1.0/1.0 \\
X-axis orientation control   & 1.0/0.4 & 1.0/0.2 & 1.0/1.0 & 0.6/1.0 & 0.6/0.0 & 0.6/0.4 \\ 
Z-axis orientation control (on the table)     & 1.0/1.0 & 1.0/1.0 & 1.0/1.0 & 0.0/1.0 & 0.4/1.0 & 0.6/1.0 \\ 
Z-axis orientation control (in the air)   & 0.6/0.4 & 0.8/0.2 & 0.6/0.8 & 0.0/0.6 & 0.0/0.0 & 0.2/0.8 \\ 
Y-axis position control      & 0.2/0.8 & 0.2/0.4 & 0.6/1.0 & 0.8/0.8 & 0.0/0.0 & 0.2/0.8 \\ \hline 

Long-horizon Tasks           &                             &                             &                             &                             &                             &                                         \\ \hline
Cable pulling*                & 1.0/0.5 & 1.0/0.5 & 1.0/0.5 & 0.0/1.0 & 0.6/0.5 & 0.4/0.5 \\ 
Direction flipping           & 1.0/0.8 & 1.0/1.0 & 1.0/1.0 & 0.0/0.6 & 0.0/1.0 & 0.0/0.0 \\ 
U-shape bending and grasping & 0.8/0.8 & 0.0/0.6 & 0.8/0.8 & 0.0/0.0 & 0.0/0.0 & 0.0/0.0 \\ 
In-hand insertion            & 0.6/1.0 & 0.0/0.4 & 0.8/0.6 & 0.0/0.8 & 0.0/0.4 & 0.2/0.6 \\ \hline 
Average Success Rate of Short-term Primitives                           & \multicolumn{3}{c}{\textbf{0.88}/0.81}                                                         & \multicolumn{3}{c}{\textbf{0.75}/0.71}                                                                     \\ 
Average Success Rate of Long-horizon Tasks                             & \multicolumn{3}{c}{0.64/\textbf{0.70}}                                                         & \multicolumn{3}{c}{0.10/\textbf{0.45}}                                                                     \\ \hline

\end{tabular}
\end{table*}

\begin{figure*}[h]
    \centering
    \includegraphics[width=0.97\linewidth]{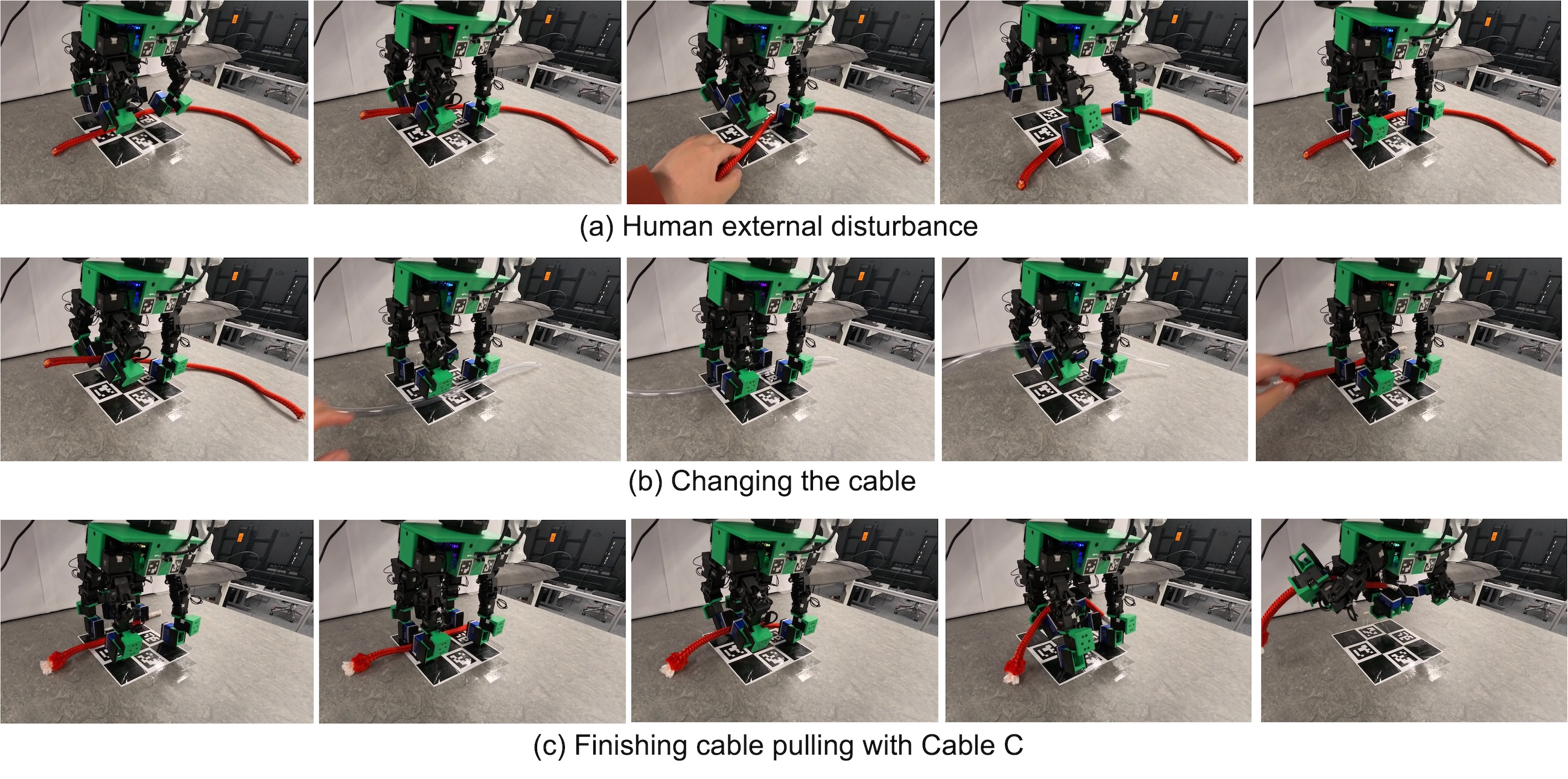}
    \caption{Human-guided primitive switching during \textbf{one} long-horizon manipulation according to the finite state machine with external intervention. (a) The grasped Cable A is dropped by a human and then the robot hand regrasps it. (b) The Cable A is replaced with Cable F, and after being VMF middle hooked, the Cable F is then replaced by Cable C again from an ungrasped pose. (c) After a pre-grasp, Cable C is successfully grasped again and pulled slowly from right to left.}
    \label{fig:exp_external_intervention}
\end{figure*}

{\bf Performance Results:} 
Each primitive was performed 5 times with different initializations on each of the 6 cable types.
We display four primitives in Figure \ref{fig:exp_primitive_demo} c to f, where the rest three primitives are intensively performed in the long-horizon manipulations that are already showned in Figure \ref{fig:intro_demo} and \ref{fig:exp_human_demonstrator}.
Table \ref{table:exp_performance} shows good performance on all 8 short-term primitives, achieving 88\% success rate of demonstration replay, when using Cables A, B, and C that were made of the same materials as the demonstration data captures (Cable A) but different diameters. 

The cables of quite different materials (Cables D, E, and F) are still manipulated well, except for the three pose control manipulations, which require very fine motor control. On the 8 short-term primitives, the demonstration replay on Cables D, E, and F achieved 75\% success rate.

The overall performance on long-horizon manipulations decreases, which is expected since each manipulation takes more than 30 seconds, and it can easily fail due to accumulated errors caused by the difference between Cable A that was used to collect demonstration data and other cables. 
For Cables A, B, and C, the demonstration replay achieved 64\% success rate on four long-horizon manipulations.
However, the success rate significantly dropped on Cables D, E, and F, which is only 10\%.

Table \ref{table:exp_performance} also compares our baseline, the human lower bound performance, for each task.
The robot hand achieved similar performance on short-term primitives compared to the lower bound of human dexterity, and was slightly worse than the lower bound on long-horizon tasks, which is expected due to the feedback control of the human hand with the limited tactile sensing.

By taking a deeper look at each primitive, we noticed the following:

1. Cables of large diameter are easier to manipulate since they are easier to grasp, comparing the success rate on Cable C (14mm) and Cable E (8mm). 
The thin cable is difficult to grasp because a small control error on the fingertips will let the cable slip from the fingertips.

2. The demonstration data was collected on a soft cable, so the experiments on stiff cables (Cable D and F) are prone to fail due to their different bending behavior during manipulation.

A straightforward solution to increase the success rate is to collect demonstration data on several cables with different physical properties, including diameters and stiffness and then replay the demonstrations on the cables with similar properties.

Besides the single primitive or long-horizon manipulation evaluation, we further evaluating the manipulation performance and the designed finite-state machine under external interventions, such as manual force and changing cables with human-guided primitive switching strategy, shown in Figure \ref{fig:exp_external_intervention}.
We used the cable pulling as an example, whose FSM is presented in Figure \ref{fig:fsm_long_horizon_fsm}.
During the manipulation, one human manually dropped the cable several times, and replaced the original cable with different cables during the manipulation.
With human-guided transitions between primitives in the FSM, the robot hand finally performed cable pulling against several interventions without failure.

\section{Conclusion}
\label{sec:conclusion}
This research is the first to propose the dexterous cable manipulation taxonomy (\textit{Cable Dexonomy}). 
We are the first to implement and evaluate a new dexterous hand with two thumbs and rotatable fingertips that is suitable for the type of dexterous cable manipulation (DCM) that \textit{Cable Dexonomy} inspires with more dexterity and better performance.
We are the first to propose a demonstration collection pipeline that allows humans to acquire training data by manipulating the hand.
Replaying this data demonstrated the ability of performing the short-term primitives and long-horizon tasks with our hand on the DCM task. 
The framework used allows the data collection to be extended, thus potentially resulting in a pipeline more widely applicable to different cable types.
Our experiments demonstrate that the collected demonstration data is robust across different unseen cables, and our hand is capable of performing tasks that are complex for humans.
Performance on long-horizon tasks based on several primitives without a prior full long-horizon demonstration also indicates that a Finite State Machine based on the primitives proposed in \textit{Cable Dexonomy} might be an approach for robust strategies for long-horizon tasks.

\section{Limitations} 
For the long-horizon experiments, we needed to manually execute the different primitives sequentially.
Without a real-time feedback, the performance of long-horizon manipulations on unseen cables is very low. 
Ongoing work is using learning-based policies for short-term primitives and investigating how to automatically switch to different short-term primitives in a long-horizon manipulation, based on observations, e.g. by using a gating neural network \cite{yang2020multi, triantafyllidis2023hybrid}.
The second limitation is that we cannot perform tasks that are extremely difficult and may requiring a moving hand rather than a fixed hand, such as making an overhand knot. 
Considering the interactions between the hand and finger motions and the arm movement is also very challenging.

\bibliographystyle{plainnat}
\bibliography{references}
\newpage

\appendices

\section{}
\label{appendix}

We introduce the supplementary content here. We also strongly encourage readers to watch our supplementary video.

\subsection{Dynamics in Cable Dexonomy}
\label{appendix:cable_dexonomy:dynamics}
Dynamics in cable manipulation occurs when the assumption breaks down that the cable's shape instantaneously follows the robot's motion, leading to inconsistencies in quasi-static modeling. 
This typically happens in cases where inertia and other dynamic effects become significant, causing a delay or deviation in the cable's response to robot motion. 
In contrast, quasi-static manipulation assumes that the cable shape continuously and smoothly follows the manipulator’s movement without significant lag or oscillation.
Most examples of tasks are quasi-static, except for cable waving \cite{chi2022iterative, zhang2021robots}, which requires a wide-range movement of the arm.
Because dynamic dexterous cable manipulation cannot be easily achieved with pure finger motions, we do not consider it in our \textit{Cable Dexonomy}.

\subsection{More Details on Long-Horizon Manipulations}
\begin{figure*}[t]
    \centering
    \includegraphics[width=0.97\linewidth]{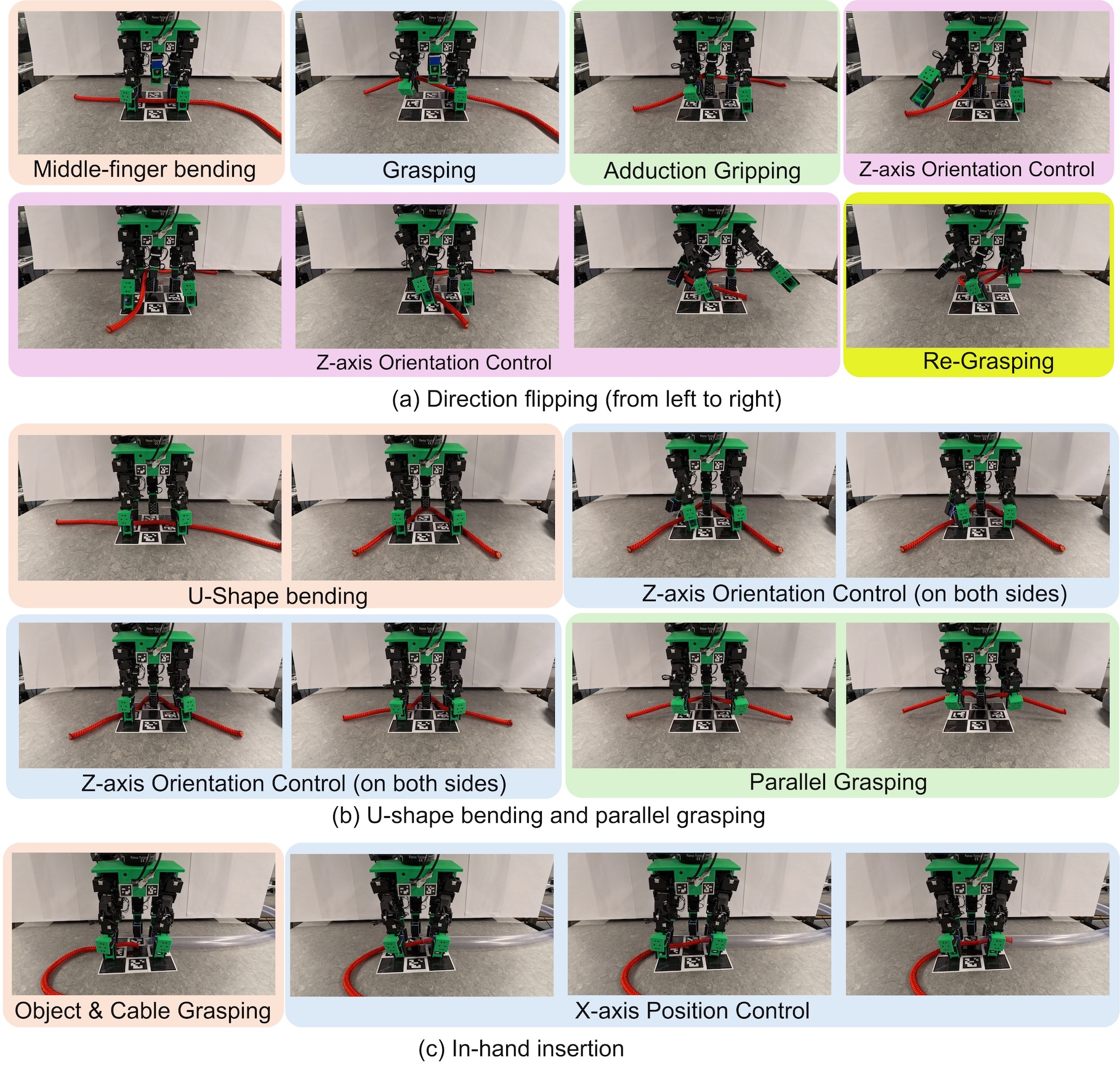}
    \caption{Demonstrations of three long-horizon manipulations. We ignore the pre-grasp which is always the first primitive to perform. (a) Direction flipping. (b) U-shape bending and parallel grasping. (c) In-hand insertion. }
    \label{fig:tax_long_horizon_task_appendix}
\end{figure*}

More demonstrations of long-horizon manipulations, including direction flipping, U-shape bending and grasping, and in-hand insertion are shown in Figure \ref{fig:tax_long_horizon_task_appendix}.

\begin{figure*}[t]
    \centering
    \includegraphics[width=0.90\linewidth]{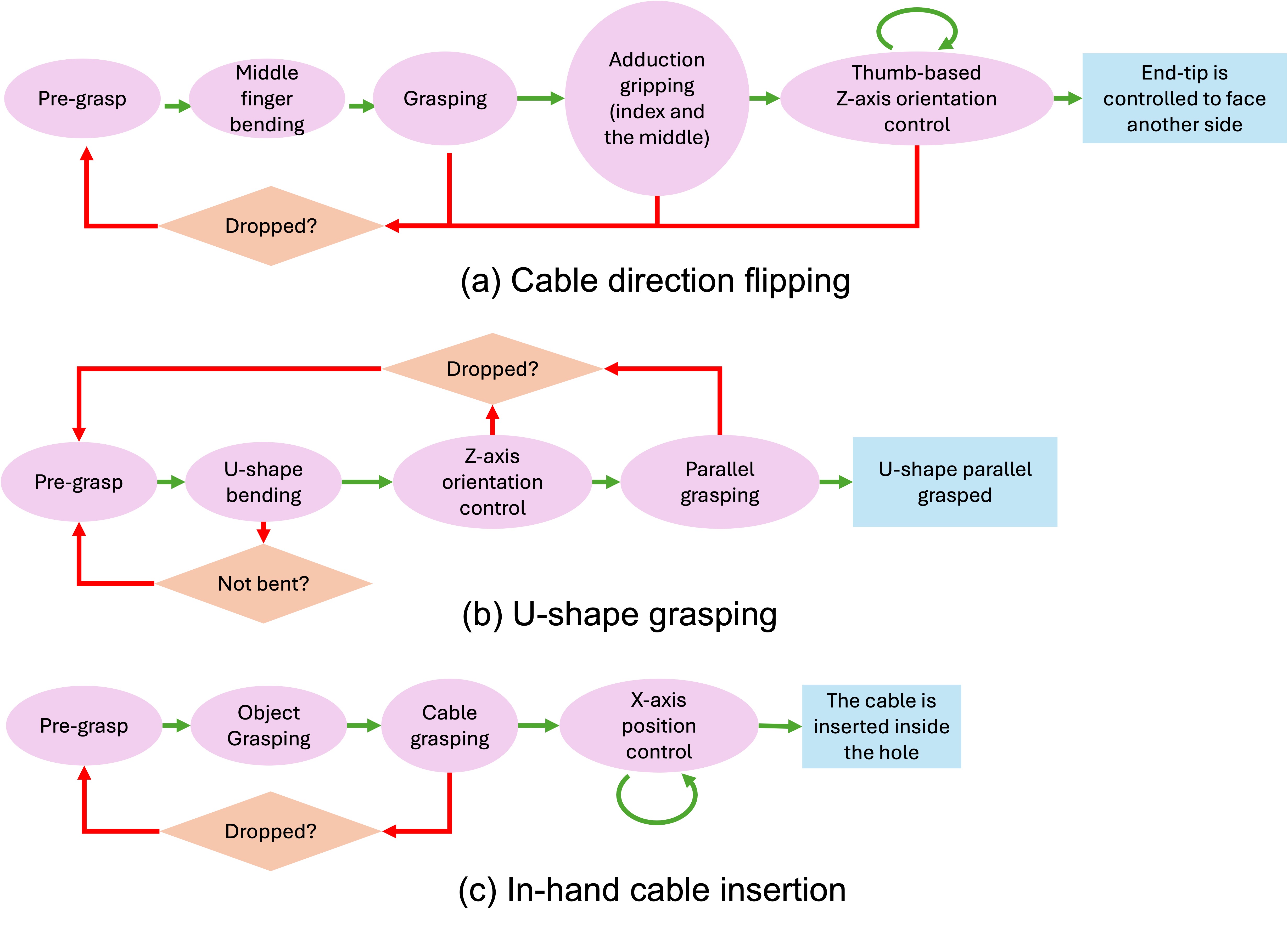}
    \caption{\textit{Cable Dexonomy} Finite State Machines of three long-horizon manipulations.}
    \label{fig:fsm_long_horizon_fsm_appendix}
\end{figure*}

More figures of finite-state machines of long-horizon manipulations performed by our designed hand are shown in Figure \ref{fig:fsm_long_horizon_fsm_appendix}.

Here are details of each long-horizon task:

1. Direction flipping: 
With the middle finger bent first, the cable needs to be grasped by the right thumb-index combo (TIC), then the middle finger and the right index finger perform adduction gripping (exactly like grasping a pencil), and the released right thumb performs Z-axis orientation control to rotate the cable's right end-tip toward the right.
Finally, the left TIC re-grasps the rotated end-tip to finish the whole direction flipping.
As mentioned in the main paper, the right thumb and the right TIC are located on the same side of the human's right hand's thumb and index finger, although the right thumb and the right TIC are on the left side of the figures.

2. U-shape bending and parallel grasping:
The middle finger first bends the cable into a U-shape, then two TICs on both sides perform Z-axis orientation control to rotate the cable's two sides, and finally two TICs perform parallel grasping.

3. In-hand insertion:
The object with the hole (the hollow transparent tube) and the cable are grasped separately by two TICs.
Then the right TIC which grasps the cable performs X-axis position control inserts the cable inside the tube repeatedly to reach a certain insertion depth.

\subsection{Minor Robot Hand Designs}
Here are additional two minor design extensions in Figure \ref{fig:hardware_minor_design_appendix}:

\begin{figure}[H]
    \centering
    \includegraphics[width=0.97\linewidth]{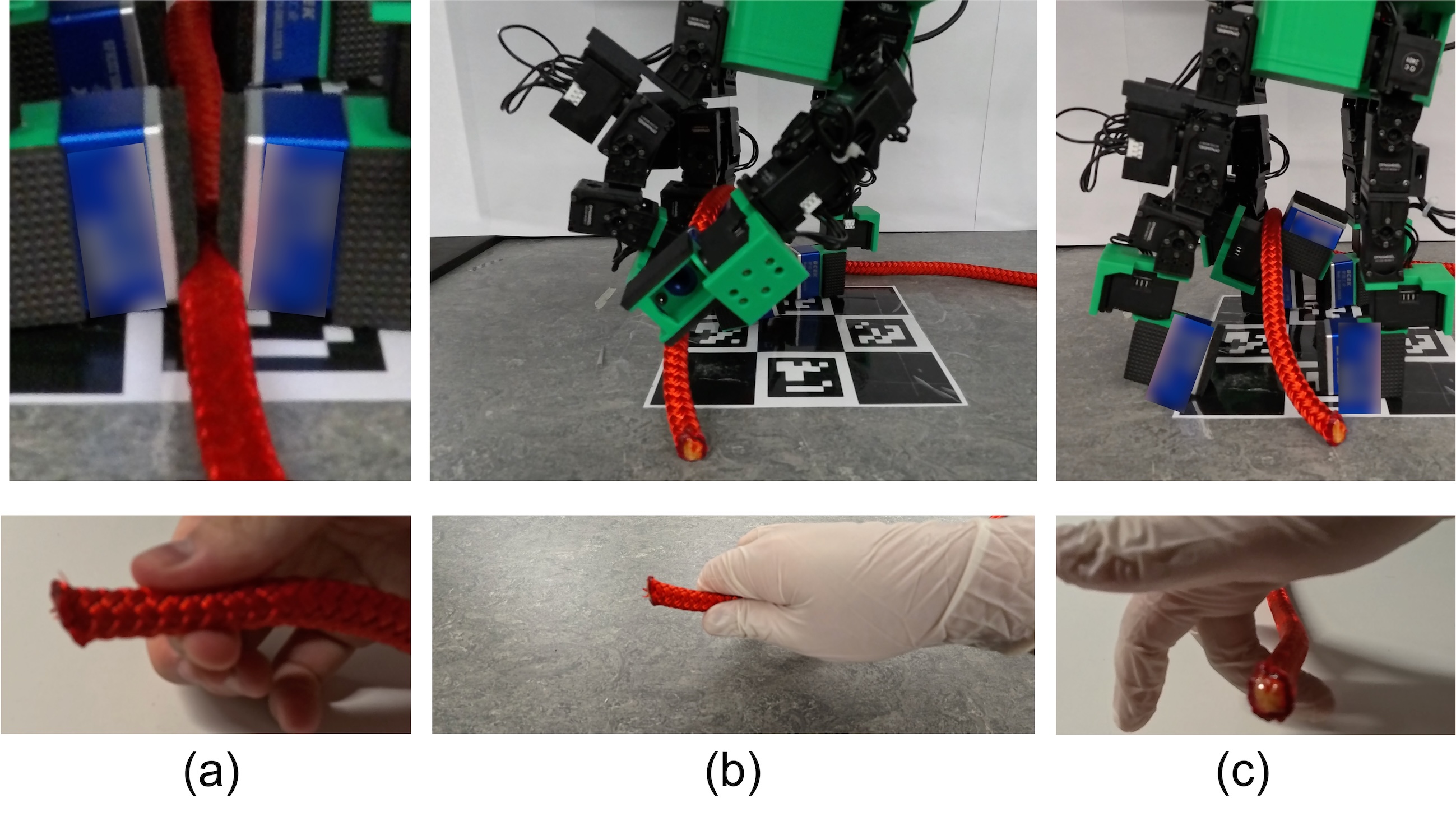}
    \caption{The top row shows some minor design details of the robot hand and the bottom raw is how the human hand behaves during cable manipulation. (a) A layer of sponge that better fits the cable. (b) Power grasp with the thumb fingertip and the middle part of the index finger. (c) Middle finger hooking with a large fingertip that can hang the cable more properly.
    }
    \label{fig:hardware_minor_design_appendix}
\end{figure}

\textbf{Attaching a thin layer of sponge on the fingertip.} This helps introduce passive compliance which acts as a mechanical buffer by mimicking human skin's deformation when grasping (Figure \ref{fig:hardware_minor_design_appendix} a). This passive compliance gives some error tolerance on fingertip positions when grasping and encourages over-grasping for larger friction and grasping stability.

\textbf{Large-size fingertips.} Instead of using flat fingertips which many anthropomorphic hands are doing, we use large fingertips. This helps two actions: 1. power grasp with the thumb's fingertip and the index finger's middle phalanx (Figure \ref{fig:hardware_minor_design_appendix} b) and 2. hook the cable on the distal interphalangeal joint (Figure \ref{fig:hardware_minor_design_appendix} c).

\end{document}